\documentclass{article}
\usepackage{graphicx}
\graphicspath{{figs/}}
\usepackage{microtype}
\usepackage{graphicx}
\usepackage{subfigure}
\usepackage{booktabs} 

\usepackage{hyperref}

\usepackage{tikz}
\usetikzlibrary{shapes.geometric, arrows}
\tikzstyle{startstop} = [rectangle, rounded corners , minimum width=1cm, minimum height=0.5cm,text centered, draw=black, fill=blue!15]
\tikzstyle{io} = [trapezium, trapezium left angle=70, trapezium right angle=110, minimum width=1cm, minimum height=0.5cm, text centered, draw=black, fill=blue!15]
\tikzstyle{process} = [rectangle, minimum width=1cm, minimum height=0.5cm, text centered, draw=black, fill=blue!15]
\tikzstyle{decision} = [diamond, minimum width=1cm, minimum height=0.5cm, text centered, draw=black, fill=blue!15]
\tikzstyle{arrow} = [thick,->,>=stealth]

\usepackage{amsmath}
\usepackage{amsfonts}
\usepackage{xcolor}
\usepackage{mdframed}
\usepackage{color,soul}
\usepackage{placeins}

\DeclareMathOperator{\E}{\mathop{\mathbb{E}}}

\DeclareMathOperator*{\argmax}{arg\,max}
\DeclareMathOperator{\GP}{\mathop{\mathcal{GP}}}

\usepackage{algorithm}
\usepackage[accepted]{icml2018}

\icmltitlerunning{Optimization, fast and slow: optimally switching between local and Bayesian optimization}

\begin{document}

\twocolumn[
\icmltitle{Optimization, fast and slow: optimally switching between local and Bayesian optimization}



\icmlsetsymbol{equal}{*}

\begin{icmlauthorlist}
\icmlauthor{Mark McLeod}{to}
\icmlauthor{Michael A. Osborne}{to,goo}
\icmlauthor{Stephen J. Roberts}{to,goo}

\end{icmlauthorlist}

\icmlaffiliation{to}{Department of Engineering Science, University of Oxford}
\icmlaffiliation{goo}{Oxford-Man Institute of Quantitative Finance}

\icmlcorrespondingauthor{Mark McLeod}{markm@robots.ox.ac.uk}

\icmlkeywords{Machine Learning, ICML, Bayesian Optimization}

\vskip 0.3in
]



\printAffiliationsAndNotice{\icmlEqualContribution} 

\begin{abstract}
We develop the first Bayesian Optimization algorithm, BLOSSOM, which selects between multiple alternative acquisition functions and traditional local optimization at each step. This is combined with a novel stopping condition based on expected regret. This pairing allows us to obtain the best characteristics of both local and Bayesian optimization, making efficient use of function evaluations while yielding superior convergence to the global minimum on a selection of optimization problems, and also halting optimization once a principled and intuitive stopping condition has been fulfilled. 
\end{abstract}
\section{Introduction}

\subsection{Bayesian Optimization with Gaussian Processes}
In Bayesian Optimization we are concerned with the global optimization of a black box function. This function is considered to be expensive to evaluate, and we are therefore willing to undertake considerable additional computation in order to achieve efficient use of evaluations. Bayesian Optimization has been applied to may problems in machine learning, including hyperparameter tuning \cite{snoek_practical_2012, hernandez-lobato_predictive_2014}, sensor set selection \cite{garnett_bayesian_2010} and tuning robot gait parameters \cite{calandra_bayesian_2016, lizotte_automatic_2007, tesch_using_2011}. A recent review of the field is \citet{shahriari_taking_2016}.

To achieve this aim at each iteration we first train a model of the objective conditioned on the data observed so far. This model is usually a Gaussian Process, a kernel-based model with particularly useful properties. A full introduction to the Gaussian Process is given by \citet{rasmussen_gaussian_2006-1}. However, the relevant properties to this work are: that all posteriors produced by the GP are multivariate Gaussian, so provide both the estimated value and the uncertainty of that estimate; and that this joint Gaussian property also extends to derivatives of the function being modelled.

We next define an acquisition function over the optimization domain which states how useful we expect an evaluation at that location to be. This function is optimized to find the location predicted to be most useful. The true objective is then evaluated at this location. There are a large number of acquisition functions available. Two that are relevant to this work are Expected Improvement \cite{jones_efficient_1998}, in which we choose the point with the greatest improvement in expectation on the best value observed so far, and Predictive Entropy Search (PES) \cite{hernandez-lobato_predictive_2014}, in which we choose the location expected to yield the greatest change in the information content of the distribution over our belief about the location of the global minimum. 

We contribute a novel algorithm, Bayesian and Local Optimisation Sample-wise Switching Optimisation Method, BLOSSOM\footnote{https://github.com/markm541374/gpbo}
, which combines the desirable properties of both local and Bayesian optimization by selecting from multiple acquisition functions at each iteration. We retain the evaluation-efficient characteristics of Bayesian optimization, while also obtaining the superior convergence of local optimization. This is combined with a Bayesian stopping criterion allowing optimization to be terminated once a specified tolerance on expected regret has been achieved, removing the need to pre-specify a fixed budget.

\subsection{Requirement for a Stopping Criterion}

In the majority of work on Bayesian optimization the problems considered either  fix the number of iterations or, less often, fix a computational budget. While this is clearly desirable for averaging over and comparing multiple repetitions of the same optimization, it is not desirable in practice: the number of steps to take (or budget to allow) is now an additional parameter that must be selected manually. This choice requires expert knowledge of Bayesian Optimization to select a number that hopefully will neither be too small, resulting in easy improvement being neglected, or too large, costing additional expensive evaluations for minimal gain. We are therefore motivated to seek an automatic stopping criterion.

Early stopping has been considered by \citet{lorenz_stopping_2015} in an application of Bayesian Optimization to brain imaging experiments. They propose and test early stopping based on the Euclidean distance between consecutive evaluations of the objective, and based on the probability of improvement on the incumbent solution. Both of these criteria provide notable improvement on a fixed number of iterations. However, both these criteria are strictly \emph{local} quantities with no consideration of the GP model at locations removed from the incumbent solution and proposed next location. 
The values must still be selected by the user so that optimization is not terminated undesirably early by an incremental exploitative step while regions of the optimization domain remain unexplored. We would prefer a stopping criterion which takes account of the full model of the objective, and which has a parameter more easily interpreted in terms of the expected difference between the proposed and true solutions.


\subsection{Convergence Properties}
\begin{figure}
\centering
\includegraphics[width=0.49\columnwidth]{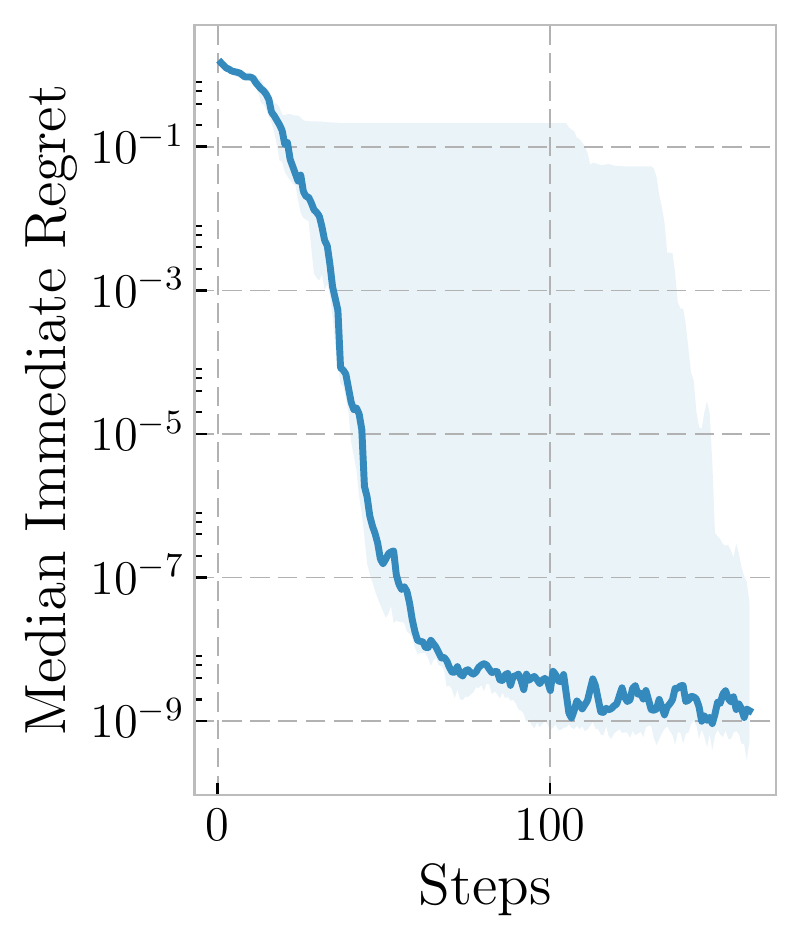}
\includegraphics[width=0.49\columnwidth]{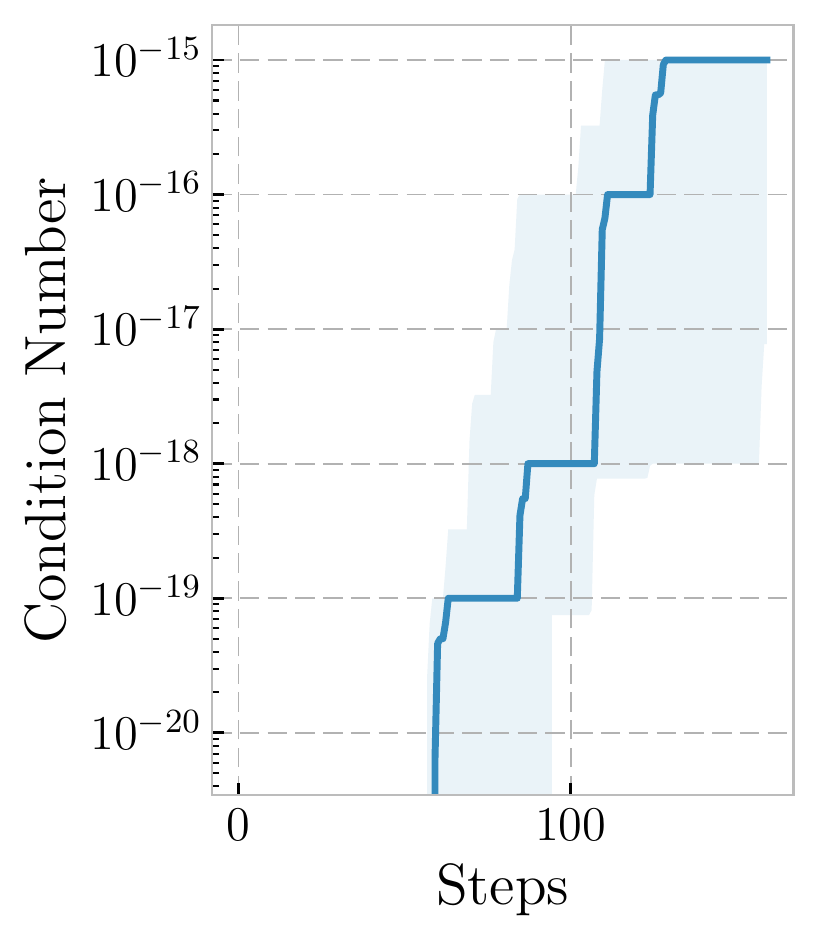}
\caption[True immediate regret]{True immediate regret and conditioning number of the covariance matrix under optimization of the Hartman-4D function using the Predictive Entropy Search acquisition function. The median and quartiles of 16 repetitions of the optimization are shown. Rather than adding a fixed diagonal component to the GP kernel matrix to ensure stability we have added the minimum value which still allows the Cholesky decomposition to succeed, in decade increments starting from $10^{-20}$. The optimization achieves a good result quickly, but then as jitter is added there is no further improvement beyond roughly $\sqrt{\text{jitter}}\,$.}
\label{jitterdemo}
\end{figure}

Optimization has excellent empirical performance in identifying the minimizer of multi-modal functions with only a small number of evaluations. \citet{bull_convergence_2011-1} has shown that $\mathcal{O}(n^{-\frac{v}{d}})$ convergence, where $v$ is the smoothness of the kernel, can be achieved with a modified version of Expected Improvement. However the authors note that this is only applicable for fixed hyperparameters. We are not aware of any estimates on convergence for PES, which exhibits better performance empirically . Furthermore, even given a guarantee of convergence in theory, details of the implementation of Bayesian Optimization ensure that the final regret is unlikely to fall to less than a few orders of magnitude below the scale of the objective function.
Firstly, we are not able to exactly maximize the acquisition function. Constraints placed on the number of evaluations available to the inner optimizer limit our ability to select evaluation points to a high degree of accuracy. This limit is also relevant to minimizing the posterior for our final recommendation.
Secondly, even in a noiseless setting, we must add diagonal noise to our covariance matrix to resolve numerical errors in performing the Cholesky decomposition. This reduces the rate of convergence available to that of optimizing an objective with the noise level we have now implicitly imposed. As we cluster more evaluations closely around the minimum, the conditioning of the covariance matrix degrades further. The potential loss in performance due to diagonal noise is illustrated in Figure \ref{jitterdemo}. We therefore also desire to create an optimization routine which does not suffer from this ineffective exploitation property. 

This convergence issue has been addressed by \citet{dhaenens_learning_2015} who switch from Bayesian Optimization to CMA-ES once a criteria on probability of improvement has been achieved. This provides excellent convergence on the objectives tested. However the switching relies on a heuristic based on a combination of previous values of the acquisition function and the number of steps without improvement of the incumbent. We would prefer to make use of a Bayesian criterion for any changes in behaviour, to avoid the need for additional parameters which must be manually chosen by an expert.
By using a non-stationary kernel which replaces the squared exponential form with a quadratic kernel in regions around local minima, \citet{wabersich_advancing_2016} also aim to achieve superior convergence. However, they do not show the final value achieved by their method in most experiments, and the use of fixed pre-trained hyperparameter samples makes their implementation unsuitable for an online setting.

We now develop our algorithm, which achieves superior convergence and terminates once a well-defined condition has been achieved. In \S \ref{sec:algorithm} we outline the behaviour of the algorithm, which selects between multiple acquisition functions on each iteration. We detail in \S \ref{sec:estimating} how we approximate the numerical quantities required, then in \S \ref{sec:localresults} provide results demonstrating the effectiveness of our new method.

\section{The Algorithm}
\label{sec:algorithm}
\subsection{Separating Global and Local Regret}
In Bayesian Optimization we aim to minimize the difference between the function value at the final recommended point, $\hat{y} = f(\hat{x})$ and the value at the true global minimizer $y_* = f(x_*)$. This is the \emph{regret} of selecting the current recommendation as the final solution. We shall now separate this concept into two distinct components which we treat separately. Let $S$ be some region of note containing the incumbent solution. We define $y_i$ as the minimum value of the objective within $S$ and $y_o$ as the minimum value outside $S$. We can then write

\begin{equation}
\begin{aligned}
\text{Regret} &= \E (\hat{y}-y_*)\\
          &= \E (\hat{y}-y_i) + \E (y_i-y_*)\\
           &= \E (\hat{y}-y_i) + \E( \max(y_i-y_o,0) )\\
          &= R_{\text{local}} + R_{\text{global}} \,,
\end{aligned}
\end{equation}

where we have split the full regret into a local component, due to the difference between our candidate point and the associated local minimum, and a global component, due to the difference between the local and global minima. Both components are non-negative by definition, and we expect both to decrease as we learn about the objective. The local component represents the difference between our incumbent and the minimum within $S$. It is reduced by exploitation of the objective. The global component represents the difference between the local minimum and the global minimum. It will be reduced by exploration, and also by exploitation of other local minima. There is a finite probability that $R_\text{{global}}=0$, corresponding to the probability that the global minimum is in fact the minimum of the basin containing our incumbent.

\subsection{Multiple Acquisition Functions}

To address the issues identified above, we split our optimization into four distinct modes, intending to use the most effective at each iteration. The modes are; Random Initialization, Bayesian Optimization, Global Regret Reduction and Local Optimization. 

Random Initialization is, as usual, only required for the first few iterations. Bayesian Optimization using Predictive Entropy Search is our default acquisition function if no relevant conditions to change behaviour have been satisfied: PES provides the usual balance between exploration and exploitation.

In steps when a distinct candidate minimum can be identified, we switch to the predominantly explorative strategy of Global Regret Reduction, intended to reduce the global regret. By making this change, we avoid the inefficient convergence  of exploitation due to poor conditioning in Gaussian Processes model when used for Bayesian Optimization. To identify a candidate minimum we require the existence of a region surrounding the minimum of the $\GP$ posterior with a high probability of being convex.

Once the predicted global regret has fallen below some target value we use a purely exploitative Local Optimization algorithm. In this work, we assume that we have access to noiseless evaluations of the objective functions so that we can employ a quasi-Newton local optimization routine, such as BFGS \cite{nocedal_numerical_2006}, which delivers super-linear convergence to the local minimum and is free of the numerical conditioning problems present in Gaussian Processes.  We note that since we are starting our local optimization very close to the minimum (in fact we choose to start only when the $\GP$ model predicts a convex region), only a small number of steps should be needed to achieve any required local regret target. We are then able to stop optimization, having achieved a target total expected regret.


 We now give further detail on the two new modes of optimization used by BLOSSOM. The methods used share many expensive computations with PES, so by reusing these results we do not incur too large an additional overhead.

\subsubsection{Global Regret Reduction}
Once a region around the posterior minimum has been identified within which local optimizations are likely to converge to the same location we do not wish to perform exploitation within this region with Gaussian Processes, as this leads to numerical conditioning issues and therefore does not use evaluations efficiently. Instead we wish to set this region aside for pure exploitation under a local search strategy. We therefore direct our efforts towards reducing the probability of any other local minima which might take lower values than our incumbent solution existing, reduction of the global regret. We use a modified form of expected improvement to achieve this, where instead of taking improvement with respect to the lowest observed objective value we compare to the estimated value of $y_i$ that would be obtained by starting a local optimization from the current incumbent. The acquisition function used is therefore
\begin{equation}
\alpha_{\text{GRR}} = (\mathbb{E}(y_i)-\mu) \Phi \left( \frac{\mathbb{E}(y_i)-\mu}{\sigma}\right) + \sigma \phi \left( \frac{\mathbb{E}(y_i)-\mu}{\sigma}\right)
\label{eq:grraq}
\end{equation}
where $\mu$ and $\sigma^2$ are the $\GP$ posterior mean and variance, $y_i$ is the minimum value within a defined region around the posterior minimum and $\Phi$ and $\phi$ are the unit Normal cdf and pdf respectively.

\subsubsection{Local Optimization}
Once we have a both sufficiently high certainty that the $\GP$ posterior minimum is close to a minimum of the objective, and that that minimum is in fact global, we wish to exploit fully.

We use the BFGS algorithm for local optimization. This is a second order algorithm which updates an estimate of the Hessian at each step. By using our estimate of the Hessian available from the GP model as the initial estimate in BFGS, we hope to achieve convergence with fewer evaluations of the objective than otherwise. Rather than modifying the BFGS algorithm to use this estimate, we rescale the problem so that the expectation of the Hessian is the identity matrix. With a local function model
\begin{equation}
f(x) = \frac{1}{2} x^T H x + x^T g + c
\end{equation}
we define $z = Rx$ where $R^{-T}=C$, $H=CC^T$ the Cholesky decomposition of $H$. This gives us a modified function
\begin{equation}
g(z) = \frac{1}{2} z^T z + z^T R^T g + c
\end{equation}
as required. Once we have started this process we are no longer performing Bayesian Optimization and can continue using BFGS until convergence. By selecting an appropriate stopping condition for local optimization we can ensure $R_{\text{local}}$ falls below any desired target. We have selected a gradient estimate of less that $10^{-6}$ as our stopping condition, but any other method could be used.

\subsection{Switching Between Acquisition Functions}

\begin{figure}
\vskip -0cm
\centering
\begin{tikzpicture}[node distance=0.8cm]
\tikzstyle{every node}=[font=\small]
\node (start) [startstop] {Random Initialization};
\node (pro1) [process, below of=start] {Train GP Model};
\node (dec1) [decision, below of=pro1, text width=1.1cm, yshift=-1.cm] {P.ve Definite at $x_{\text{min}}$};

\node (dec2) [decision, right of=dec1, text width=1.3cm, xshift=4.2cm,yshift=-0.8cm] { $R_{\text{global}}\leq \text{Target}$};
\node (pro3) [process, above of=dec2, yshift=1.2cm, text width=2cm] {Estimate $R_{\text{global}}$};
\node (pro2) [process, above of=pro3, text width=2cm ,yshift=0.4cm] {Estimate P.ve Region Radius};

\node (pro4) [process, below of=dec1, yshift=-1.3cm, text width=2cm] {Maximize PES Acquisition};
\node (pro5) [process, below of=dec2, yshift=-0.3cm, xshift=-2.4cm,text width=2cm] {Maximize Global Regret Acquisition};

\node (in1) [io, below of=pro4, text width=3cm, yshift=-0.1cm] {Evaluate Objective};
\node (stop) [startstop,below of=dec2,xshift=0.cm,yshift=-1.4cm] {Local Exploitation};
\node (tmp) [left of=pro2,xshift=-2.25cm,yshift=-1cm]{};
\node (tmp4) [below of=tmp]{};
\node (tmp2) [left of=dec1,xshift=-1cm]{};
\node (tmp3) [below of=tmp2,yshift=-0cm]{};
\draw [arrow] (start) -- (pro1);
\draw [arrow] (pro1) -- (dec1);
\draw [arrow] (dec1) -- node[anchor=east]{no} (pro4);
\draw [arrow] (pro4) -- (in1);
\draw [arrow] (pro2) -- (pro3);
\draw [arrow] (pro3) -- (dec2);
\draw [arrow] (dec2) -| node[anchor=west,yshift=0.3cm]{no}(pro5);
\draw [arrow] (dec2) -- node[anchor=east,yshift=0cm]{yes}(stop);
\draw [arrow] (dec1) -| node[anchor=north,xshift=-0.2cm]{yes} (tmp);
\draw [arrow] (tmp4) |- (pro2);
\draw [arrow] (pro5) |- (in1);
\draw [arrow] (in1) -| (tmp2);
\draw [arrow] (tmp3) |- (pro1);
\end{tikzpicture}
\caption[Flowchart for acquisition function switching behaviour.]{Flowchart for acquisition function switching behavior. Following initialization the acquisition function may switch between PES and Regret Reduction until $R_{\text{Global}}$ achieves a sufficiently low value. The final steps are then under the local exploitation strategy.}
\label{fig:aqswitchflowchart}
\vskip 0cm
\end{figure}
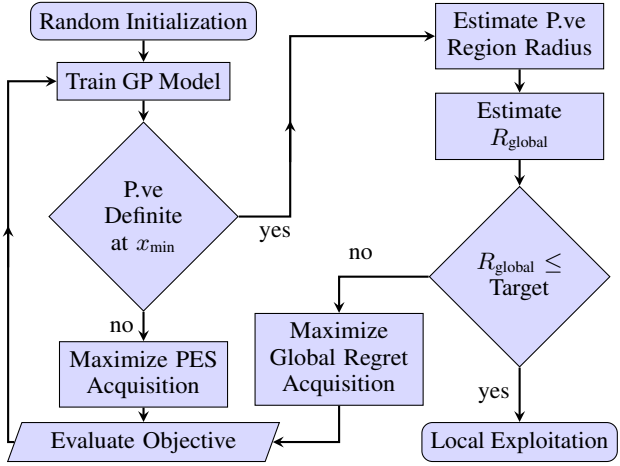

We have specified that we wish to make decisions on the basis of the existence of a candidate minimum, and on the value of the global regret. In Figure \ref{fig:aqswitchflowchart} we show the decision making process used. 
However, we have not yet specified these criteria exactly.
We choose to consider the existence of a sphere with non-zero radius, centred on the minimum of the GP posterior mean, $x_{\text{min}}$, within which the objective function has a high probability of being convex at all points. Local optimization routines have excellent performance on convex functions, so if our model predicts a convex region surrounding a local minimum with high confidence we no longer desire our Bayesian Optimization to recommend inefficient exploitative evaluations in this region. We therefore switch to the Global Regret Reduction acquisition function, which will place a high weight on exploration. 

\begin{figure}
\vskip 0cm
\centering
\includegraphics[width= 0.95\columnwidth]{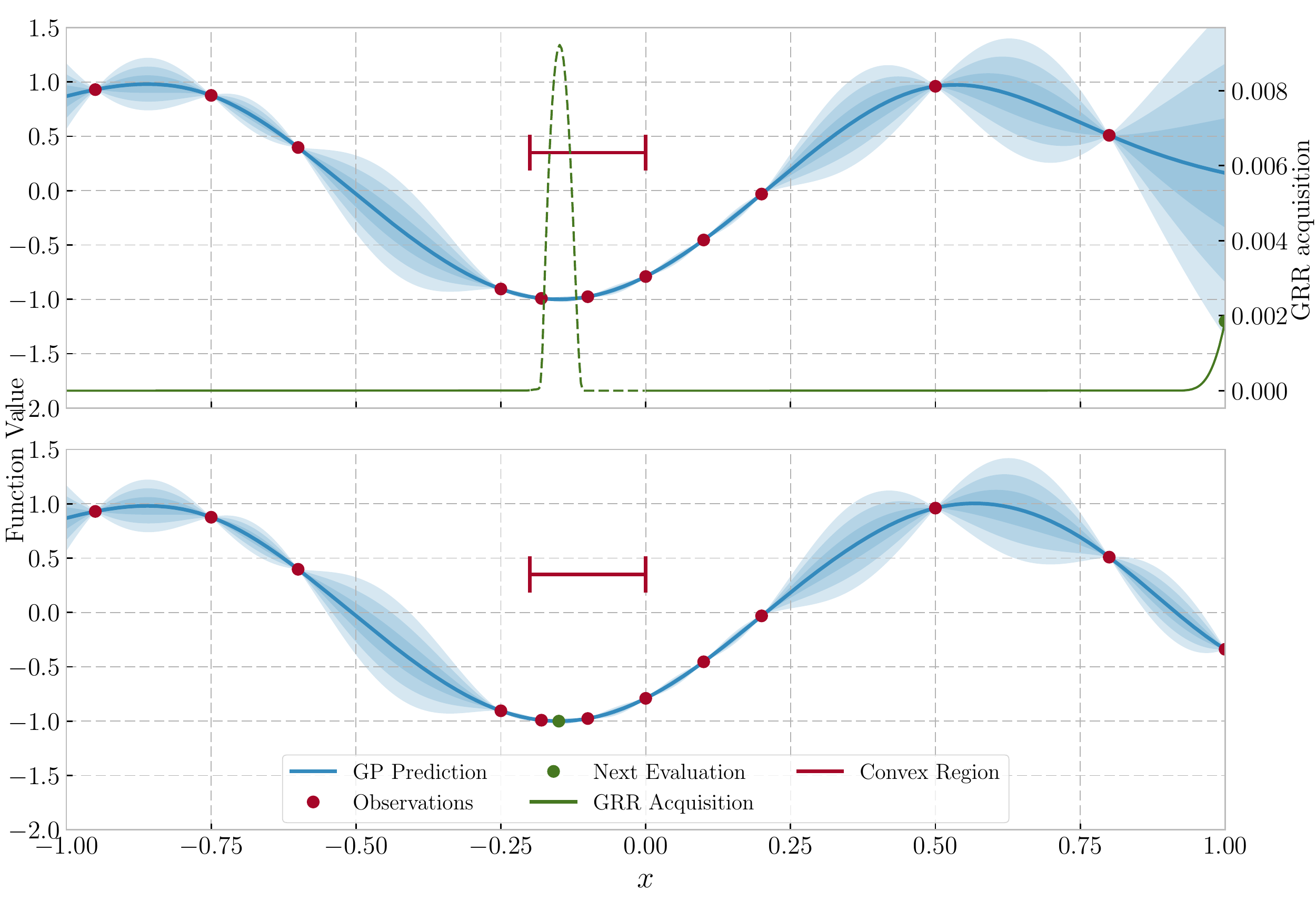}

\caption{Two steps of BLOSSOM in an example problem. In the upper plot the GRR acquisition has a maximum at -0.15. However, since this falls within the convex region around the minimum of the posterior mean the next function evaluation is taken at the highest location outside this region. In the lower plot the estimated Global Regret is sufficiently low that no further Bayesian iterations are required. The next evaluation is the start of a local optimization from -0.15.}
\label{fig:cartoon}
\vskip -0.25cm
\end{figure}

We have defined the global regret as the difference between the objective value at the local minimizer in some region and the true global minimum. We choose to use the positive definite sphere to define this region. We can then obtain an estimate of Global Regret. If this estimate falls below our target value we move to the final stage of optimization and use Local Exploitation, otherwise we continue with the Global Regret Reduction. This process is illustrated in Figure \ref{fig:cartoon}

\section{Estimating Required Quantities}
\label{sec:estimating}
We now detail the procedures used to estimate the quantities required in our switching criteria. We first detail our method of determining a positive definite region, then provide a method for estimating the global regret and expected local minimum value.
\subsection{Identifying a Convex Region}
Convexity is characterized by the Hessian matrix of the objective being positive definite at all points.  For a matrix $H$ to be positive definite we require 
\begin{equation}
x^T Hx > 0
\label{eq:pvedef}
\end{equation}
 for all $x$. Given a Gaussian Process model of the objective we would like to construct:
\begin{enumerate}
\item{A method to determine, using our GP model, the probability that the Hessian is positive definite at any point; and}
\item{A method to determine, using our GP model, the largest region centred on the current posterior minimum with a required probability of being convex at all points within that region.}
\end{enumerate}

\subsubsection{Convexity at a Point}

We make use of the Cholesky decomposition to determine if a matrix is positive definite. A unique real solution to the Cholesky decomposition of a matrix only exists if the matrix is positive definite.  Implementations of the routine commonly return an error rather than computing the complex solution. We can make use of this behaviour to provide a test for positive definiteness returning a binary result: $D(X) : \mathbb{R}^\frac{d(d+1)}{2} \rightarrow \{0, 1\} $ where $d$ is the dimensionality of the problem.

Since under a Gaussian Process model there will always be non-zero probability over all real values of inferred quantities we can never have certainty of positive definiteness.
We therefore wish to determine the probability that the Hessian of our objective at some point $x$ is positive definite (or if the point is on the boundary of the domain the Hessian for the remaining dimensions) under our GP model. All elements of the Hessian have a joint Normal distribution $H \mid x,M \sim \mathcal{N} \big( H_\mu(x), H_\sigma(x)\big)$ with mean and variance given by the GP posterior at $x$ (only elements of the upper triangle need to be included in implementation since the Hessian is symmetric). The probability of positive definiteness at $x$ is then
\begin{equation}
\begin{aligned}
p\big(D(H) \mid x,M\big) &= \int p\big(D(H) \mid H\big)p\big(H \mid x,M\big) dH \\
  & = \int D(H) p\big(H \mid x,M\big) dH \\
  &= \lim_{N \to \infty} \frac{1}{N} \sum_{i=1}^{N} D(h_i)
\end{aligned}
\end{equation}

where $M$ is our GP model and the $h_i$ have been drawn from the multivariate normal $p(H \mid x,M)$.



As our test of positive definiteness at a point we require all of some $n$ samples of $H$ to be positive definite. We treat the positive definiteness of samples from $p(H \mid x,M)$ as Bernoulli distributed with rate parameter $\theta$, since it is a deterministic binary output function of $H$. 
Taking a uniform prior on $\theta$ the 
 posterior expected value of $\theta$ is
\begin{equation}
\mathbb{E}[\theta] =  \int_0^1 \theta p(\theta) d \theta = \frac{n+1}{n+2} \,.
\end{equation}
Passing our test for positive definiteness at a point, as described in Algorithm \ref{alg:pdtest}, can therefore be interpreted as determining that $\mathbb{E}(\theta)=1-\epsilon$ where $\epsilon = \frac{1}{n+2}$ while failure implies $\mathbb{E}(\theta) < 1-\epsilon$.

\begin{algorithm}
\begin{algorithmic}
\STATE {\bfseries Input:}  location $x$, tolerance $\epsilon$
 \STATE $\texttt{G}\gets \texttt{GP\_model}$
 \STATE $\texttt{Hmean}, \texttt{Hvar}\gets \texttt{G.infer\_Hessian}(x)$
 \STATE $\texttt{PVEcount} \gets 0$
 \STATE $n \gets \tfrac{1}{\epsilon}-2$
\FOR {$i = 1 ... n$}
    \STATE $h \gets \texttt{draw\_Gaussian}(\texttt{Hmean}, \texttt{Hvar})$
    \STATE $h^* \gets \texttt{remove\_boundary\_elements}(h)$
    \IF {$\texttt{Cholesky}(h^*) \neq \text{FAIL}$}
    \STATE $\texttt{PVEcount} \gets \texttt{PVEcount} + 1$
    \ENDIF
\ENDFOR
\STATE $p \gets \frac{PVE\text{count} + 1}{n+2}$
\STATE  $\textbf{Return}\quad p \geq 1-\epsilon$

\end{algorithmic}
\caption{Positive Definite Test}
\label{alg:pdtest}
\end{algorithm}

\subsubsection{Radius of a Convex Region}
\label{sec:convexrad}
The method above allows us to effectively test a point for convexity. We now wish to use this function to find a convex region centred around the posterior minimum (again we exclude any axes on the boundary of the search domain). We choose to find the hypersphere centred at $x_{\text{min}}$ with the greatest possible radius $R_{\text{max}}$. As before we can not obtain a certain value. Instead we find an estimate, $\hat{R}_{\text{max}}$, which is the minimum distance to a non-positive definite  over a finite set of test directions $u$.

We draw unit vectors, $u$, uniformly at random, by normalizing draws from the multivariate normal distribution $u = \frac{v}{|v|}$ where $v \sim \mathcal{N}(0,I_{d})$. For each direction we obtain the positive definite radius
\begin{equation}
R_{u}(u) = \argmax_{PD(\hat{x}+ru)=1 } r
\end{equation}
by performing a binary linesearch on $r$ down to a resolution $h_r$. The first search is performed with the radius of the search domain as the upper limit, subsequent directions use the previous value of $R(u)$ as the upper limit and test the outer point first, moving on to the next direction if this point returns a convex result. We thus obtain
\begin{equation}
\hat{R}_{max} = \min_{u \in U} R_u(u)
\end{equation}
which is in the minimum distance from $\hat{x}$ to the edge of the positive definite region out of $n_u = \| U \|$ random directions as our estimate of the radius of a convex spherical region centred on $x_{\text{min}}$.
\begin{algorithm}
\caption{Positive Definite Sphere Radius}
\begin{algorithmic}
\STATE {\bfseries Input:}  center $x_{\text{min}}$, number of directions, $n_u$ tolerance $\epsilon$
\STATE $u\gets \texttt{random\_unit\_vector}$
\STATE $x_{\text{edge}} \gets \texttt{dist\_to\_domain\_boundary} $
\STATE $\hat{R} \gets \| x_{\text{min}} -x_{\text{edge}} \|$
\FOR {$i = 1 ... n_u$}
	\IF {$D(x+\hat{R}u) =0$}
        \STATE $\hat{R}\gets \texttt{binarysearch}(u,\hat{R})$
    \ENDIF
    \STATE $u\gets \texttt{random\_unit\_vector}$
\ENDFOR
\STATE $\textbf{Return}  \: \hat{R}$

\end{algorithmic}
\end{algorithm}

To obtain an estimate of the global regret we must marginalize over the values of the local and global minima, $y_o$ and $y_i$. We assume independence between these quantities, a reasonable assumption since alternative locations for the global minimizer are usually in separate basins to the incumbent. The expectation is therefore
\begin{equation}
R_{global} =  \iint_{y_i \times y_o} \max(y_i - y_o,0)\, p(y_i ) p(y_o) \,dy_i\,dy_o
\end{equation}
If we consider $y_i$ to be well approximated by a Normal distribution $\mathcal{N}(\mu_i, \sigma_i^2)$ then we can perform the integral over $y_i$
\vskip -0.5cm
\begin{equation}
\begin{aligned}
R_{\text{global}} &= \int_{y_o} \int_{y_i=y_o}^{+\infty} \max(y_i - y_o,0) p(y_i) \,dy_i \,p(y_o) dy_o\\
& \mkern-32mu = \int_{y_o} \bigg[(\mu_i-y_o) \Phi \left( \frac{\mu_i-y_o}{\sigma_i} \right) \\
& \qquad \qquad \qquad \qquad +\sigma_i \phi \left( \frac{\mu_i-y_o}{\sigma_i} \right) \bigg] p(y_o) dy_o \\
& \mkern-32mu \approx  \sum_{j}^N (\mu_i-y_o^{(j)}) \Phi \left( \frac{\mu_i-y_o^{(j)}}{\sigma_i} \right)+\sigma_i \phi \left( \frac{\mu_i-y_o^{(j)}}{\sigma_i} \right) \,.
\end{aligned}
\label{eq:regretsummation}
\end{equation}
Since we do not have an analytic form for $p(y_o)$ we are not able to perform the second integral. We instead approximate the marginalization over global regret as a summation.

To evaluate this expression we can draw $N$ samples from our GP model and find the value of $y_o$ in each case. This cannot be performed exactly, and instead we must take draws from the GP posterior over some set of support points $X_s$. Half of these are approximately drawn from the distribution of the global minimum using the method described by \citet{mcleod_practical_2017} (slice sampling over the Expected  Improvement or LCB as suggested by \citet{hennig_entropy_2012} could equivalently be used), while half of them are drawn using rejection  sampling with the GP posterior variance as an unnormalized distribution, to provide additional support in high variance regions outside the convex region.
To evaluate $\mu_i$ and $\sigma_i$ we can use the same set of draws, considering this time only points within the convex region, to obtain a sequence of samples of $y_i$ which can be used for a maximum likelihood estimate of the mean and variance of a normal distribution.
\section{Results}
We compare BLOSSOM to Expected improvement with the PI stopping criteria of \citet{lorenz_stopping_2015}, and to PES using the acquisition function value as the stopping criterion. For each algorithm we test multiple values of the stopping criteria, shown in the legend as appropriate.  

\label{sec:localresults}
\subsection{In-Model Objectives}
\begin{figure*}
\centering
\includegraphics[width=0.385\textwidth]{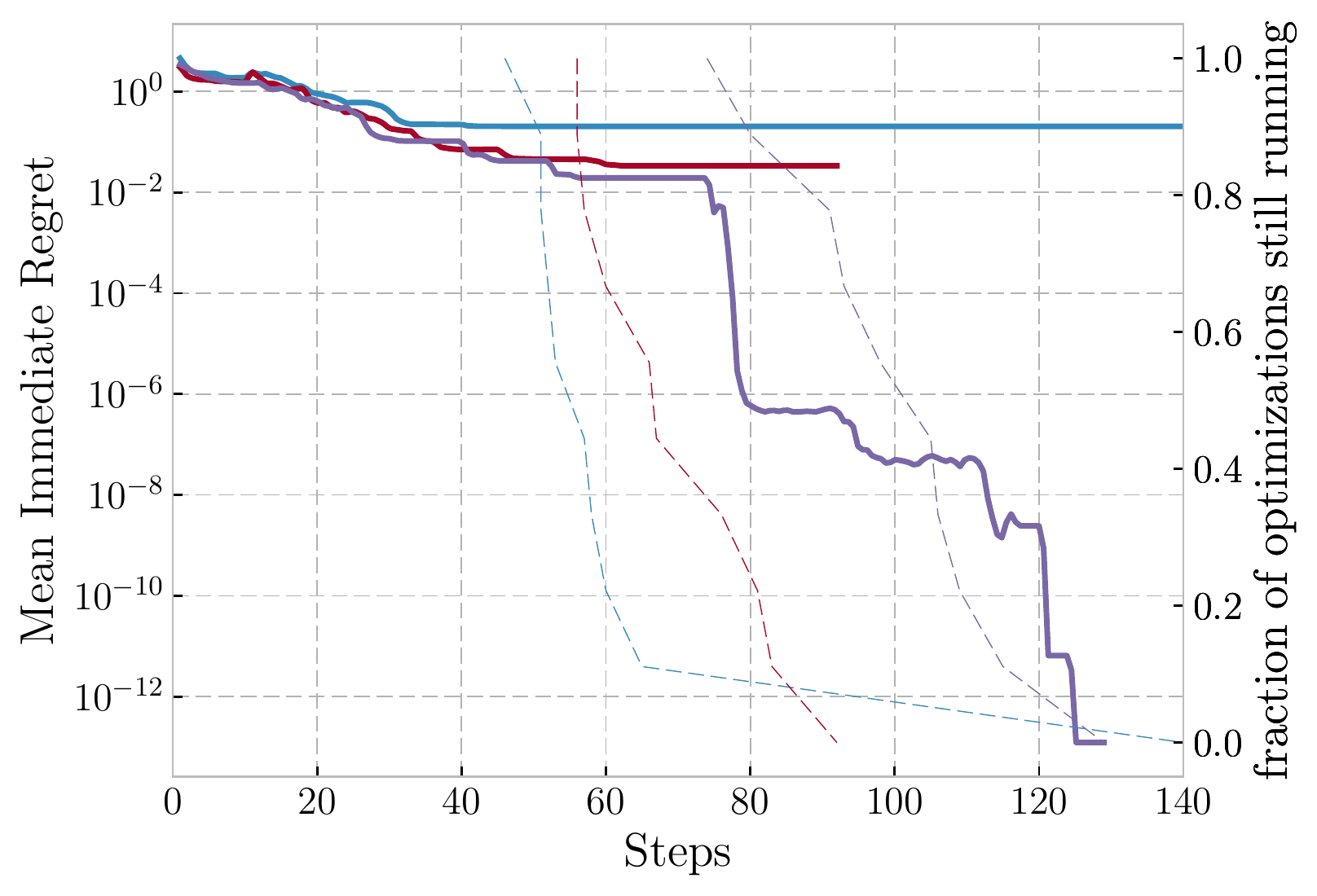}
\includegraphics[width=0.546\textwidth]{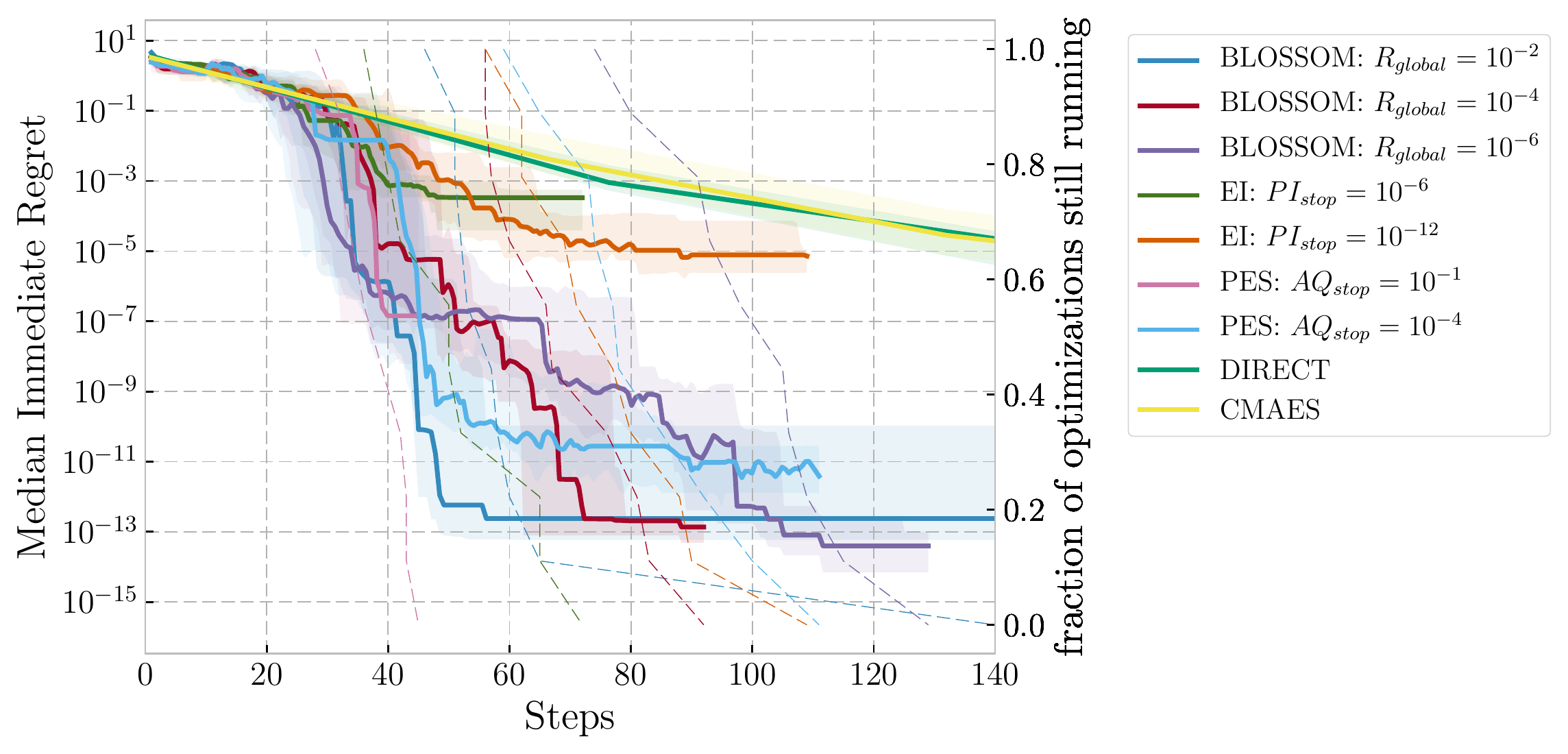}
\caption[Global regret on draws from kernel]{Comparison of methods on Draws from the Mat\'ern 5/2 kernel in 2D (left), and illustration of the effect of changing our stopping parameter on the mean expected regret (right). Mean regret is heavily influenced by a few large values. Out of 35 runs there are those targeting  $10^{-2}$ regret had 6 non-trivial results, targeting $10^{-4}$ yielded 4 non-trivial results, targeting  $10^{-6}$ achieved the global minimum on every occasion while with no stopping condition only one failure occurred. Although with a limited number of the mean regret is not particularly close to the target values the decrease with more stringent stopping conditions is clear. We have also included DIRECT \cite{jones_lipschitzian_1993} and CMA-ES \cite{doi:10.1162/106365603321828970} to illustrate the performance on non-Bayesian methods. }
\label{BH2}
\end{figure*}

To demonstrate the effect of changing the target value of global regret we make use of objective drawn from a GP, since the effect may not be observable using any single fixed objective. For example, the Branin function has multiple equal-valued global minima. We will always achieve the global minimum, and the target regret only alters the number of steps required to terminate. We show in Figure \ref{BH2} the mean regret over objectives drawn from the Mat\'ern 5/2 kernel and note that we have achieved roughly the values we requested for expected regret.

\subsection{Common Benchmark Functions}
\begin{figure*}
\centering
\includegraphics[width=0.33\textwidth]{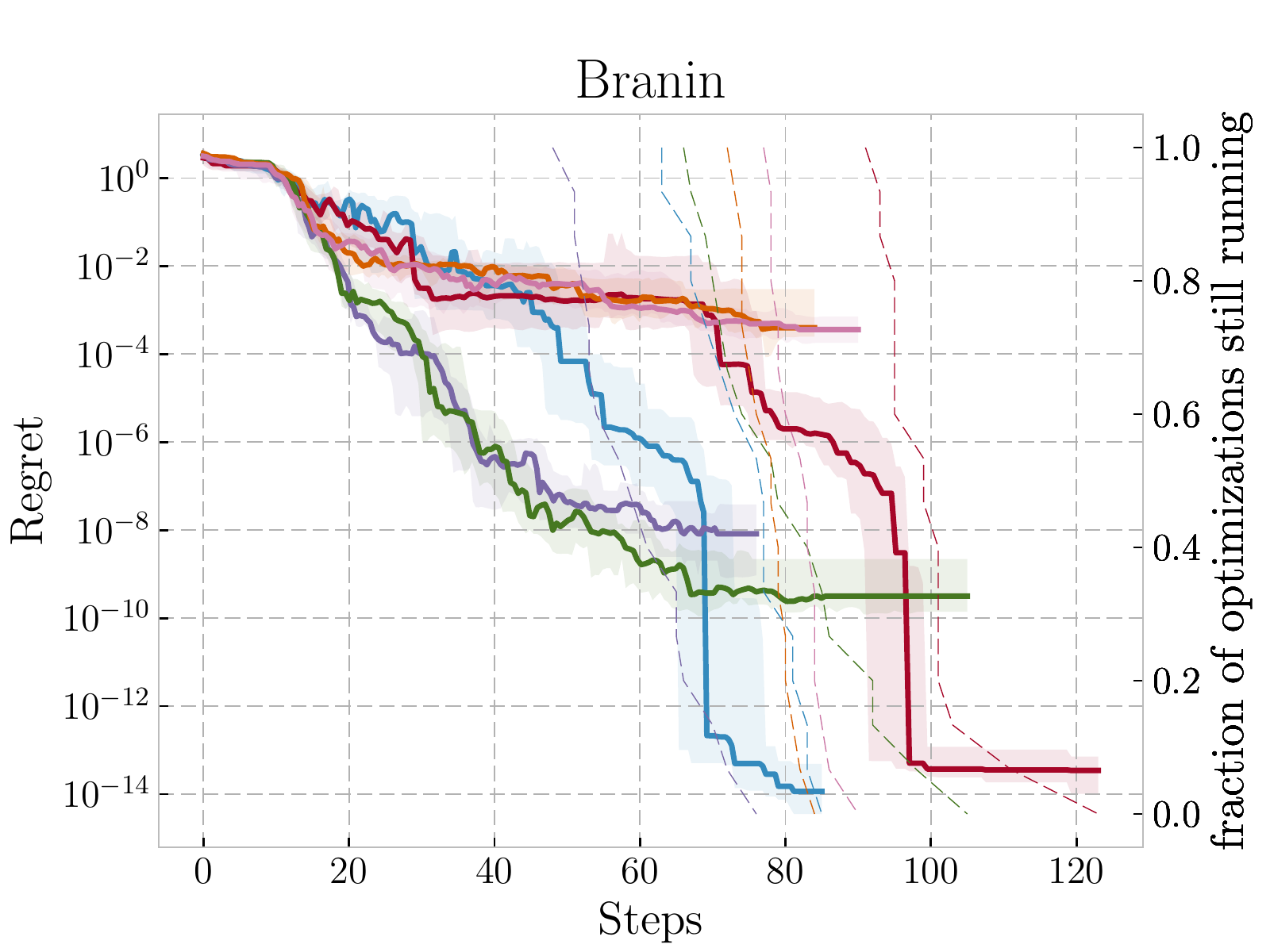}
\includegraphics[width=0.33\textwidth]{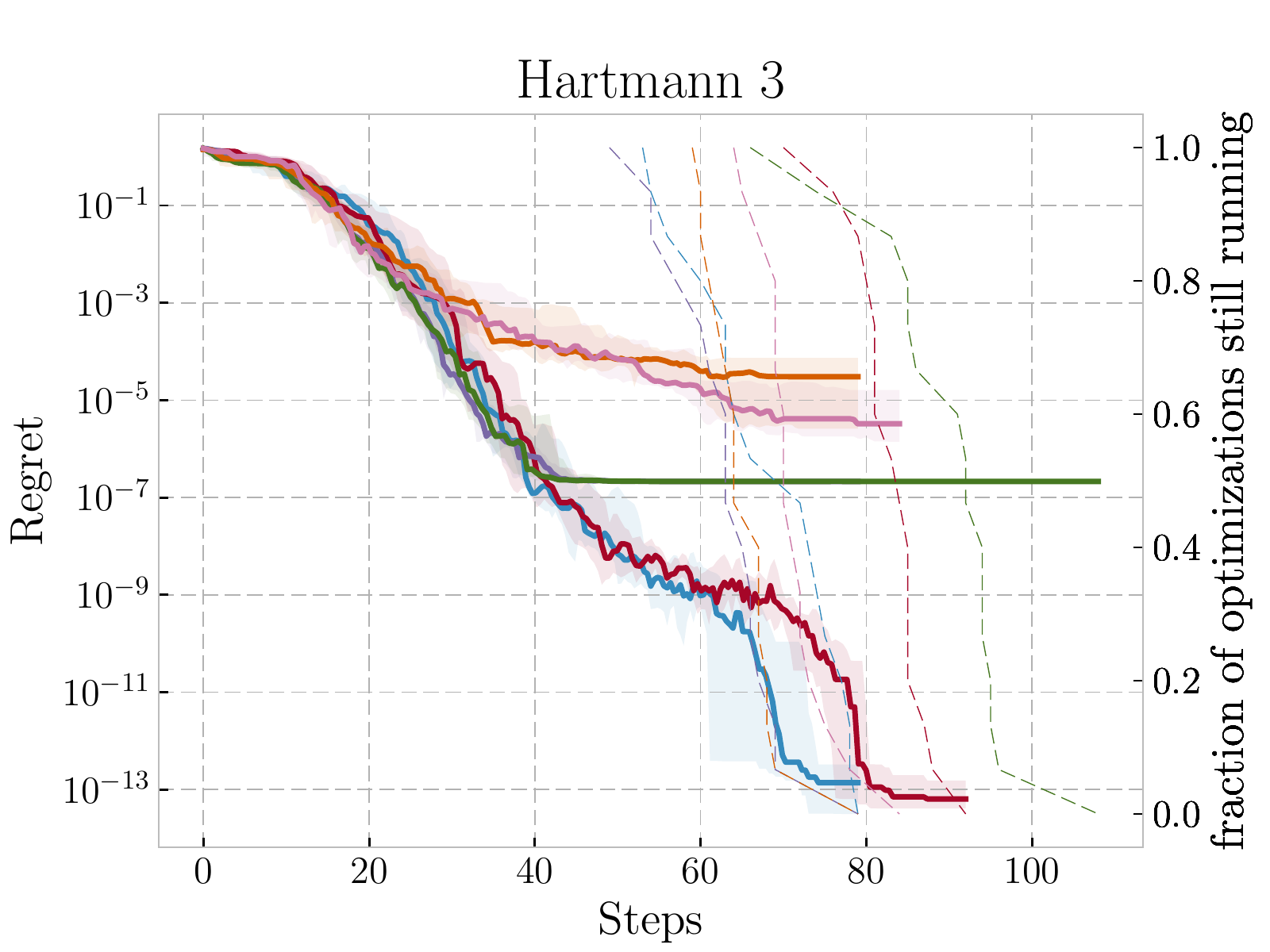}
\includegraphics[width=0.33\textwidth]{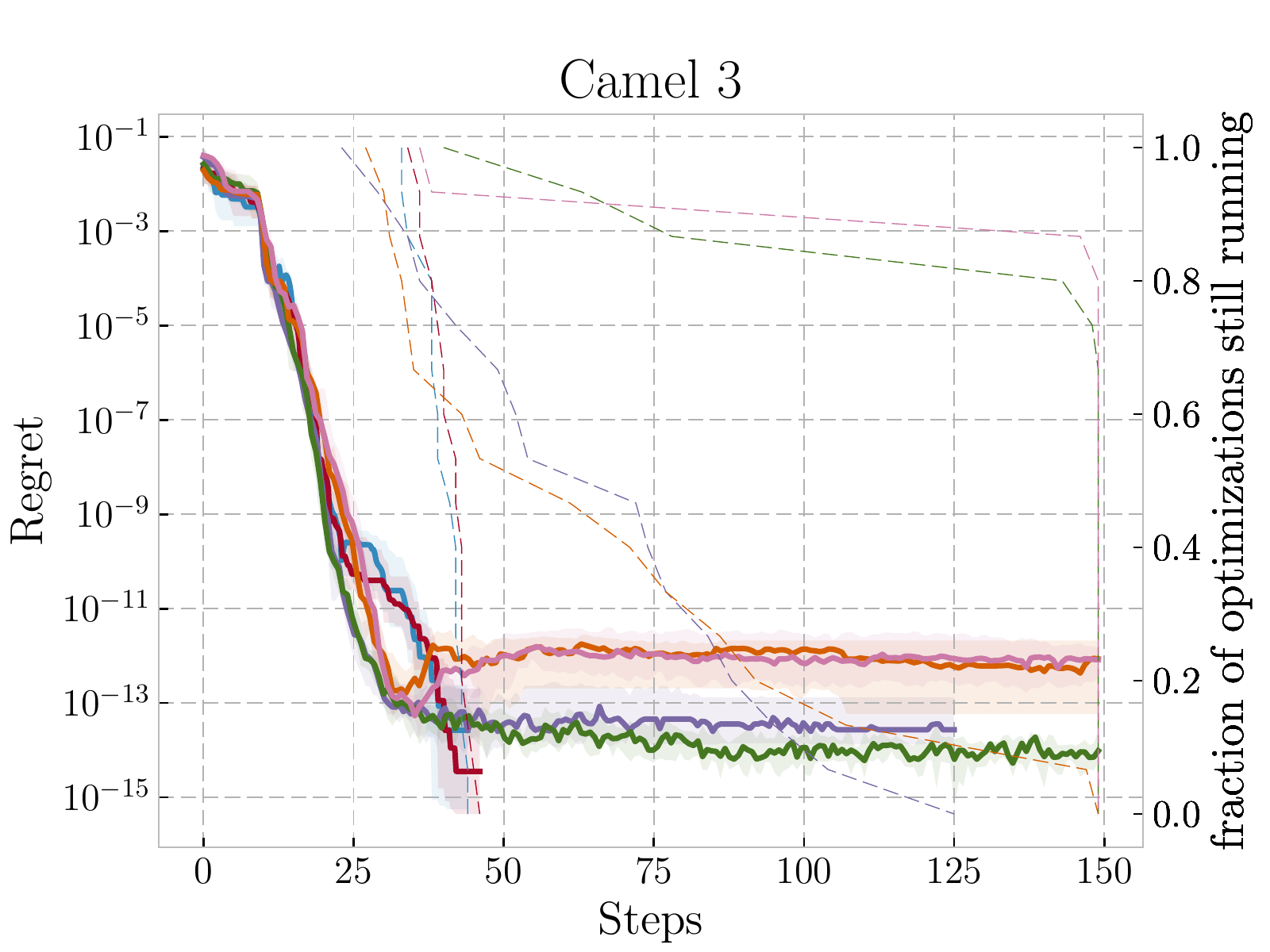}\\
\includegraphics[width=0.33\textwidth]{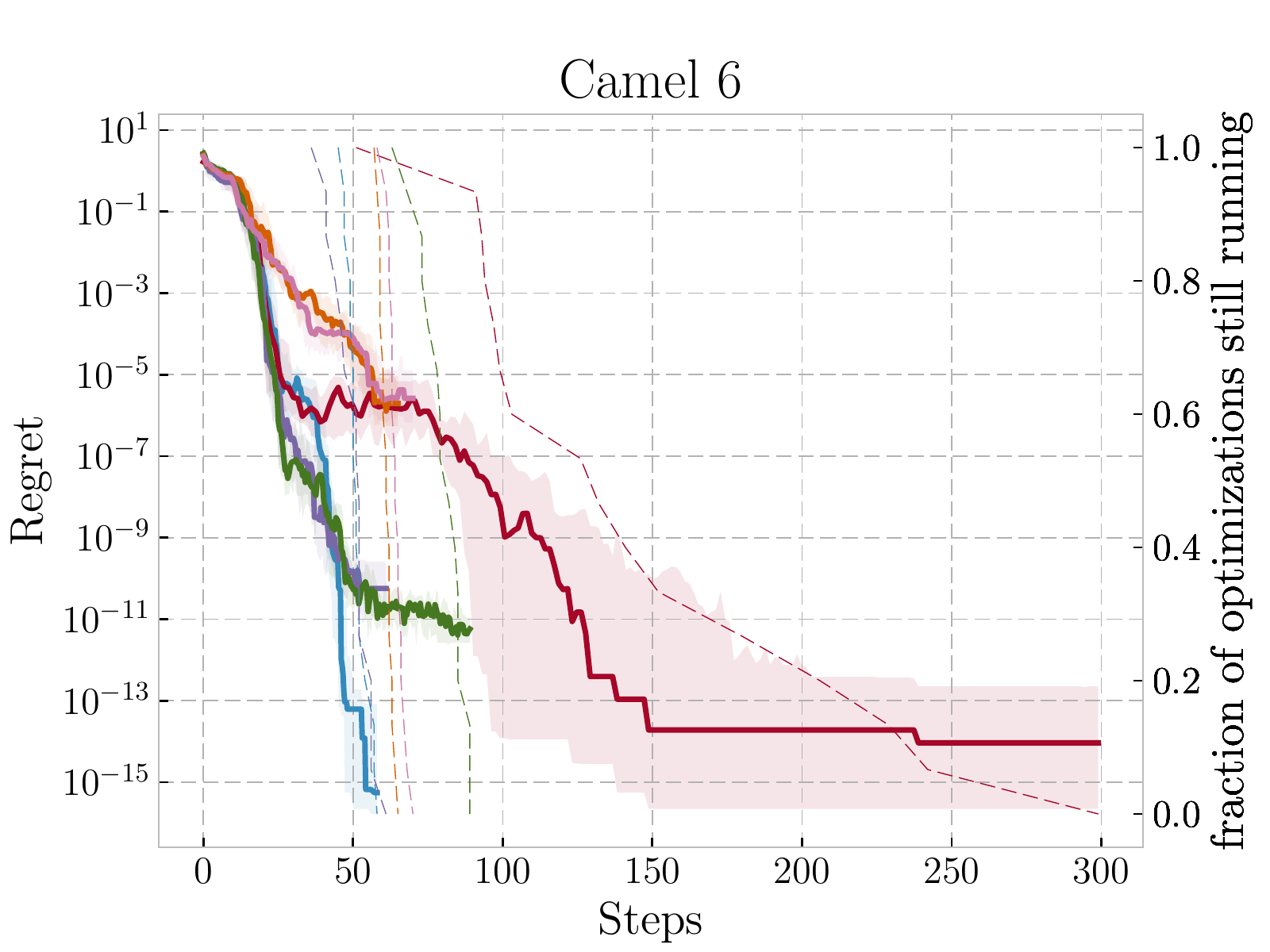}
\includegraphics[width=0.33\textwidth]{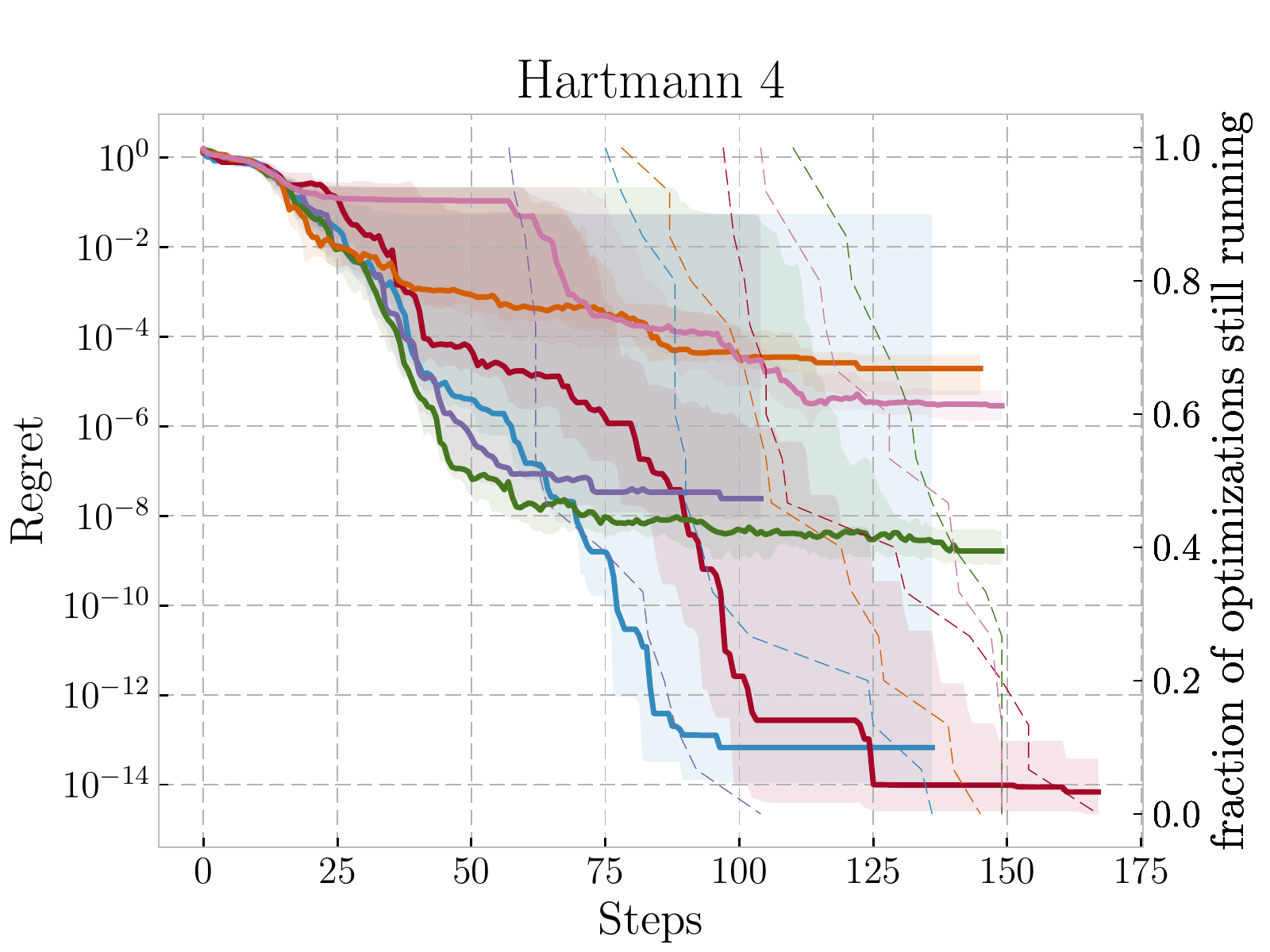}
\includegraphics[width=0.33\textwidth]{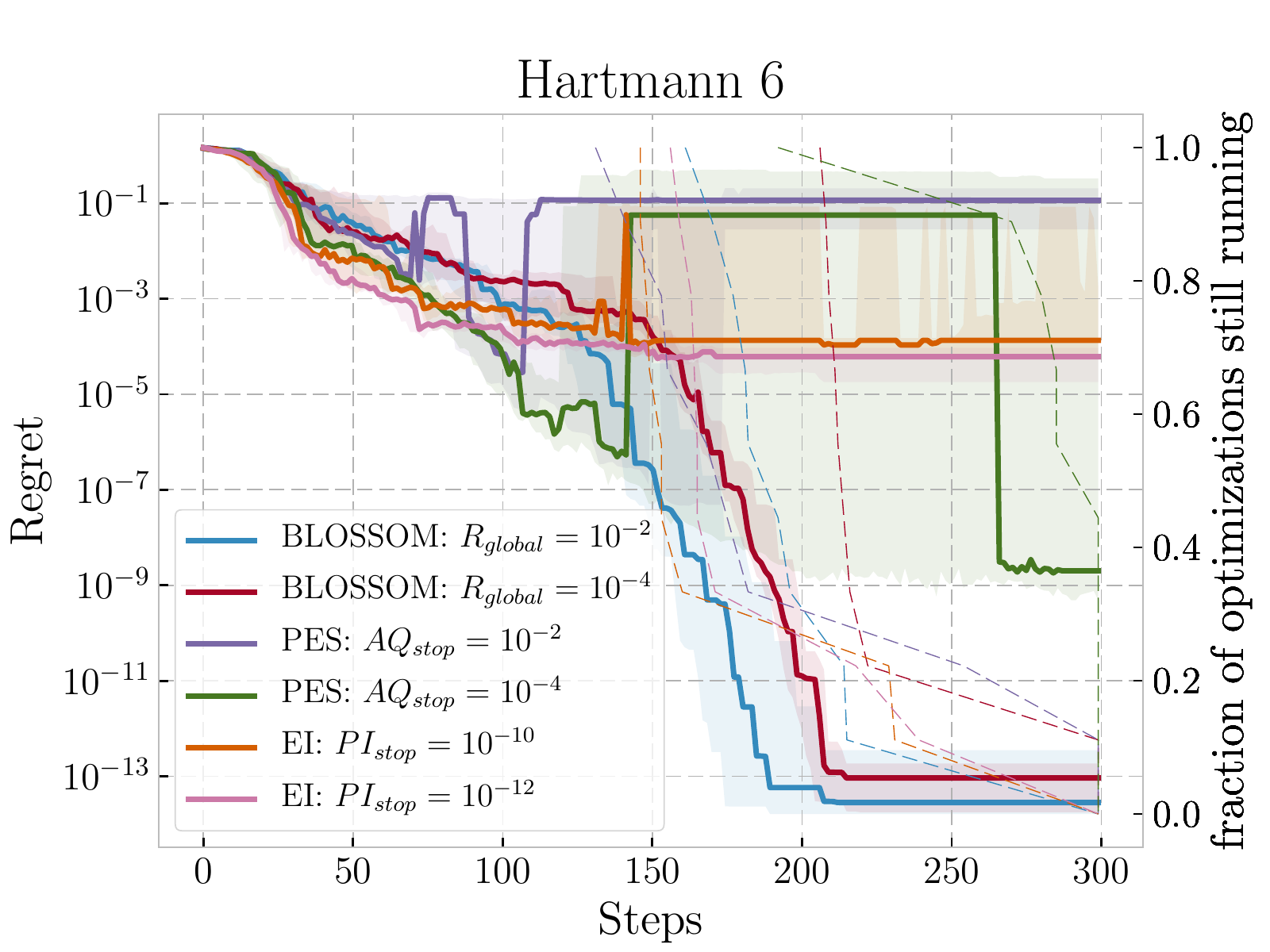}
\caption[Performance on common functions]{Comparison of various stopping methods. All stopping criteria allow an average saving compared to continuing to run optimization for many steps past convergence, but our method reliably achieves a low value before stopping. The median and quartiles of regret are shown. Fraction of optimizations still running after n steps are shown in thin dashed lines.}
\label{stopplot}
\end{figure*}

We now give results for several common test objectives for global optimization, illustrated in Figure \ref{stopplot}. In these tests we have transformed the objectives by $y' = \log(y-y_*+1)$. This is unrelated to our contributions, and is done as many of the functions used take the form of a flat plain surrounded by steep sides many orders of magnitude greater than the plain. This shape is extremely dissimilar to draws from the Mat\'ern $\frac{5}{2}$ kernel used by our GP model, so yields very poor results. This is an ad-hoc transformation, and it would be preferable to either use a kernel more appropriate to the objective or learn a transform of the output space online as suggested by \citet{snelson_warped_2004-1}. 

Neither the number of steps taken nor the regret achieved is alone a useful metric for the effectiveness of a stopping condition (few steps with high regret are obviously undesirable, but also a small decrease in regret may not be worth a much increased number of steps), so in Table \ref{stopresult} we have also shown the mean product of steps and regret, $\mathbb{E}[nR]$. Equal contours of this metric take the form of $y=\frac{a}{x}$, so low values indicate improved performance. 

As is clear from the median curves in Figure \ref{stopplot}, and mean final values in Table \ref{stopresult}, we have been successful in achieving both superior local convergence and early stopping. BLOSSOM achieves the lowest mean terminal regret, and mean product of regret and iterations, for five of the six test objectives. There is considerable disparity between the plotted and tabulated results for the Hartman 3D and 4D functions. However, we argue that this is in fact correct behaviour. These objectives are characterized by having multiple local minima of differing values. Usually Bayesian optimization will identify the correct basin as the global minimum and our local optimization converges to the correct value, as is evident in Figure \ref{stopplot}. However, with some non-zero probability the GP will predict the global minimum and its surrounding positive definite region in the wrong location. If the estimated global regret is less than our target value when this occurs, the solution is accepted, leading to exploitation of a local minimum and a high final regret. This occurs on several runs of our algorithm when using a value of $10^{-2}$ as the target global regret. When the lower target value of $10^{-4}$ is used additional exploration steps are required to reduce the global regret estimate. These provide additional opportunities to correctly identify the basin of the global minimum. This leads to the much greater reliability observed in Table \ref{stopresult} at the cost of an increased number of objective evaluations.


\begin{table*}
\begin{center}
\begin{tabular}{c c c c c c c} \toprule

Objective & BLOSSOM $10^{-2}$&  BLOSSOM $10^{-4}$ & PES $10^{-8}$ & PES $10^{-10}$ & EI $10^{-10}$ & EI $10^{-12}$\\ \midrule
&\multicolumn{6}{c}{Regret} \\ \cmidrule(r){2-7}
Branin       &\textbf{3.32e-14} & 5.2e-07 & 1.09e-07 & 2.9e-09 & 0.00221 & 0.00125\\
Camel 3hump  &2.26e-13 & 1.79e-13 & 4.14e-13 & \textbf{2.44e-14} & 2.12e-12 & 1.57e-12\\
Camel 6hump  &\textbf{2.28e-14} & 7.95e-13 & 9.41e-11 & 1.62e-11 & 2.32e-05 & 2.35e-05\\
Hartmann 3D  &0.107 & \textbf{1.14e-13} & 2.16e-07 & 2.16e-07 & 6.72e-05 & 0.000116\\
Hartmann 4D  &0.0534 & \textbf{5.21e-14} & 0.0534 & 0.0133 & 6.06e-05 & 6.44e-06\\
Hartmann 6D  &\textbf{0.00371} & 0.0638 & 0.196 & 0.157 & 0.0229 & 0.0669 \\

 &\multicolumn{6}{c}{Steps} \\ \cmidrule(r){2-7}

Branin       &74.6 & 99.8 & \textbf{59.6} & 80.4 & 77.4 & 81.9\\
Camel 3hump  &\textbf{39.6} & 40.9 & 64.9 & 132 & 66.8 & 135\\
Camel 6hump  &51.7 & 139 & \textbf{49.8} & 78.6 & 61.2 & 64.7\\
Hartmann 3D  &67.8 & 82.6 & \textbf{62.8} & 89.4 & 65.1 & 71.1\\
Hartmann 4D  &98.5 & 122 & \textbf{72.3} & 134 & 111 & 130\\
Hartmann 6D  &199 & 230 & 196 & 281 & \textbf{181} & 190\\ 

  &\multicolumn{6}{c}{Steps $\times$ Regret} \\ \cmidrule(r){2-7}

Branin       &\textbf{2.39e-12} & 5.15e-05 & 5.69e-06 & 2.22e-07 & 0.167 & 0.103 \\
Camel 3hump  &8.56e-12 & 7.02e-12 & 1.18e-11 & \textbf{2.73e-12} & 1.07e-10 & 2.31e-10 \\
Camel 6hump  &\textbf{1.14e-12} & 2.21e-10 & 4.84e-09 & 1.33e-09 & 0.00138 & 0.00157\\
Hartmann 3D  &5.98 & \textbf{9.41e-12} & 1.35e-05 & 1.93e-05 & 0.00422 & 0.00746\\
Hartmann 4D  &6.93 & \textbf{5.89e-12} & 4.36 & 1.95 & 0.00527 & 0.000853\\
Hartmann 6D  &\textbf{1.11} & 19.1 & 37.6 & 46.2 & 5.28 & 17.1\\

 \bottomrule
\end{tabular}
\end{center}
\caption{Performance of selected stopping methods on various common objective functions. For two stopping criterion values for each algorithm we show the final regret, number of steps taken and step-regret product. Our methods have achieved the best regret and step-regret product on five of the six objectives used. }
\label{stopresult}
\end{table*}

\subsection{GP Hyperparameter Optimization}
\begin{figure}[H]
\centering
\includegraphics[width=0.92\columnwidth]{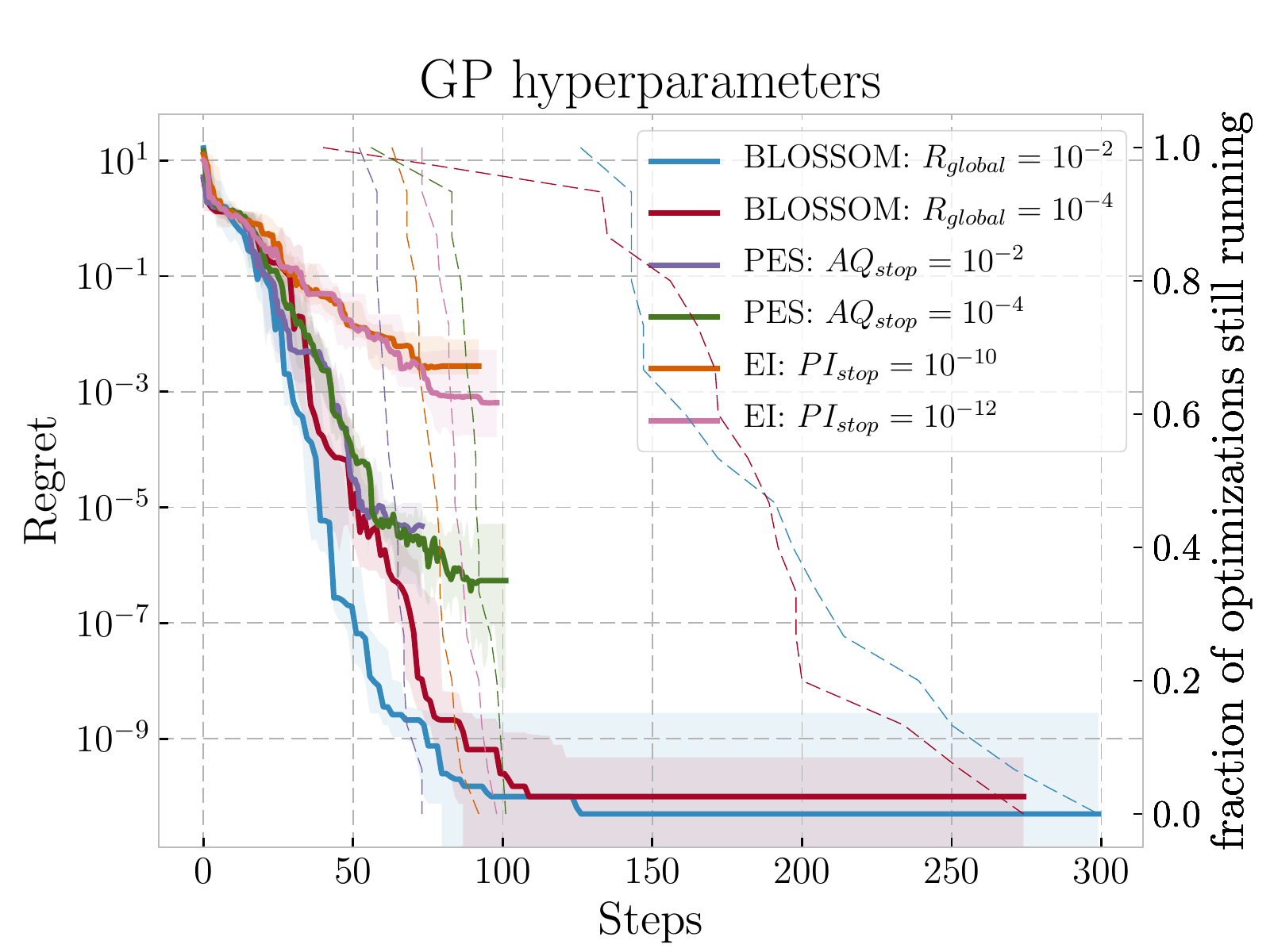}

\caption[Performance GPhyp]{Comparison of stopping criteria optimizing the hyperparameter log-likelihood of a Gaussian Process. Regret is shown with respect to the best value achieved by any single optimization, the median and quartiles of 16 repetitions of each method are shown. Our method consistently obtains several orders of magnitude better convergence than PES or EI at termination.}
\label{gphyp}
\end{figure}

Optimizing model hyperparameters is a common problem in machine learning. We use BLOSSSOM to optimize the input and output scale hyperparameters of a Gaussian Process using 6 months of half hourly measurements of UK electricity demand during 2015 \footnote{www2.nationalgrid.com/UK/Industry-information/Electricity-transmission-operational-data/Data-explorer}. As shown in Figure \ref{gphyp} we are able to obtain the best absolute result while terminating within a reasonable number of iterations, avoiding taking unnecessary further evaluations once the optimum has been achieved.
\section{Conclusion}

We have developed BLOSSOM, a Bayesian Optimization algorithm making use of multiple acquisition functions in order to separately consider exploration and exploitation by actively selecting between Bayesian and local optimization. This separation allows us to avoid the poor local convergence of Gaussian Process methods. We are further able to halt optimization once a specified value of global regret has been achieved. This has the potential to save considerable computation in comparison to manual specification of the number of iterations to perform. We have shown that BLOSSOM is able to achieve an improvement in the final result of several orders of magnitude compared to existing methods.




\FloatBarrier

\bibliographystyle{icml2018}

\end{document}